\documentclass[preprint,12pt]{elsarticle}

\usepackage{amssymb}
\usepackage{amsmath}  
\usepackage{graphicx}  
\usepackage{multirow}
\usepackage{algorithmic}
\usepackage[ruled, lined, linesnumbered, longend]{algorithm2e}
\journal{Computer Vision and Image Understanding}

\begin{document}

\begin{frontmatter}



\title{Representation Learning of Point Cloud Upsampling in Global and Local Inputs}

\author[ECUST,PolyU]{Tongxu Zhang}
\author[ECUST]{Bei Wang}
\affiliation[ECUST]{organization={East China University of Science and Technology},
             addressline={130 Meilong Road},
             city={Shanghai},
             postcode={200237},
             state={Shanghai},
             country={China}}
\affiliation[PolyU]{organization={The Hong Kong Polytechnic University},
             addressline={Hung Hom, Kowloon},
             city={Hong Kong},
             postcode={},
             state={Hong Kong SAR},
             country={China, Hong Kong}}

%\author{}

%\affiliation{organization={},%Department and Organization

\begin{abstract}
In recent years, point cloud upsampling has been widely applied in tasks such as 3D reconstruction and object recognition. This study proposed a novel framework, ReLPU, which enhances upsampling performance by explicitly learning from both global and local structural features of point clouds. Specifically, we extracted global features from uniformly segmented inputs (Average Segments) and local features from patch-based inputs of the same point cloud. These two types of features were processed through parallel autoencoders, fused, and then fed into a shared decoder for upsampling. This dual-input design improved feature completeness and cross-scale consistency, especially in sparse and noisy regions. Our framework was applied to several state-of-the-art autoencoder-based networks and validated on standard datasets. Experimental results demonstrated consistent improvements in geometric fidelity and robustness. In addition, saliency maps confirmed that parallel global-local learning significantly enhanced the interpretability and performance of point cloud upsampling.

\end{abstract}




\end{frontmatter}


\section{Introduction}
LiDAR and depth cameras saw widespread use due to advancements in 3D sensors and 3D scanners, mostly relying on point clouds to collect data. However, issues such as sparsity, uneven distribution, noise, and redundancy were common in point cloud data collection. Due to unavoidable flaws in point cloud data collection, such as occlusion, generating denser and higher-quality 3D point clouds through point cloud upsampling methods, similar to tasks performed on complete point clouds, was essential. This allowed point clouds to fully demonstrate their significant value in 3D reconstruction \cite{lin2022cosmos, peng2023deep}, object detection and classification \cite{cai2018cascade, tang2023sca, tan20233d}, and robotic operations \cite{varley2017shape,xu2022fpcc}. And in the future, it will be further applied to cloud platforms \cite{chakraborty2021secure,tiwari2022efficient}, which can be integrated with the local environment.

The irregular nature of point clouds made them easy to update \cite{chang2015shapenet}, as each point in a point cloud was independent, making it convenient to add new points and perform interpolation \cite{xia2021asfm}. However, it was challenging to apply convolution to point clouds when using learning-based methods. Nonetheless, through graph-based methods \cite{scarselli2008graph}, works such as \cite{qi2017pointnet, yuan2018pcn, wang2019dynamic} facilitated the application of deep learning in point cloud tasks. Subsequently, PU-Net \cite{yu2018pu} proposed the first deep learning-based upsampling method for point clouds. With the development of Graph Convolution Networks (GCN) \cite{henaff2015deep}, models like MPU/3PU \cite{yifan2019patch} used clustering for interpolation, enabling better local and edge recognition in point clouds. Building on this, PU-GCN \cite{qian2021pugcn} further improved feature extraction, constructing DenseGCN for extracting local input features from point clouds, achieving better results. Additionally, PU-GAN \cite{li2019pugan} applied generative adversarial networks \cite{goodfellow2020generative}, using generators to upsample and produce high-fidelity, realistic point clouds. PU-Transformer \cite{qiu2022putransformer} further advanced this field by utilizing positional information and self-attention mechanisms for enhanced global feature extraction.

Although significant progress was made in point cloud upsampling through deep learning \cite{zhao2022self, he2023gradpu, kim2024puedgeformer}, substantial challenges remained. As noted, unlike voxel-based grid representations of 3D models \cite{liu2019relation, zhou2018voxelnet}, each point in a point cloud was independent. Ergo, \cite{han2024s3u} extracted features from voxels to reduce the problem of inaccurate feature extraction caused by the irregularity and sparsity of point clouds. But, for the point clouds, independence brought unique challenges in extracting objective features of objects, particularly in incomplete or sparse regions with limited information. This was akin to the task of working with incomplete point clouds, where global features were extracted from partial inputs but lost geometric details due to incompleteness \cite{zhang2020detail}. Therefore, effectively aligning local details with global information was crucial. While Zhang \cite{zhang2024rethinking} discussed the influence of local and global inputs on point cloud upsampling, it lacked interpretability and breadth in experimental models. Current methods \cite{li2021point, du2022cascaded, he2023gradpu} utilized refiners on coarse point clouds to merge local and global features, but their networks, which were single-encoder structures, faced limitations in ensuring consistency and precise alignment of cross-scale features. BiPU \cite{zhu2024bilevel} employed asymmetric, dedicated encoders for local and global inputs, intending to combine local and global feature information efficiently through parallel extraction. Indeed, upsampling largely relied on local information and knowledgeable global priors. Parallel extraction, similar to that in image domains \cite{zhao2020representation}, was necessary, but the need for highly specialized global and local feature extraction remained in question. To avoid complexity in feature extraction within the encoder part, we proposed using the same encoder for both global and local features, thus avoiding complex networks.

\begin{figure}[tb]
    \centering
    \begin{minipage}{0.75\linewidth}
		\centerline{\includegraphics[width=\textwidth]{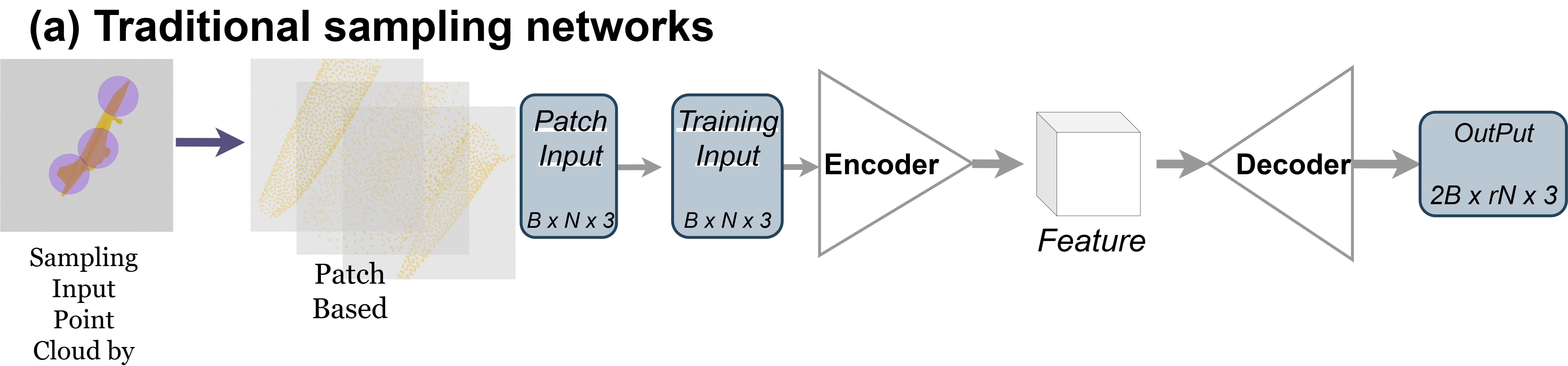}}
		\centerline{\includegraphics[width=\textwidth]{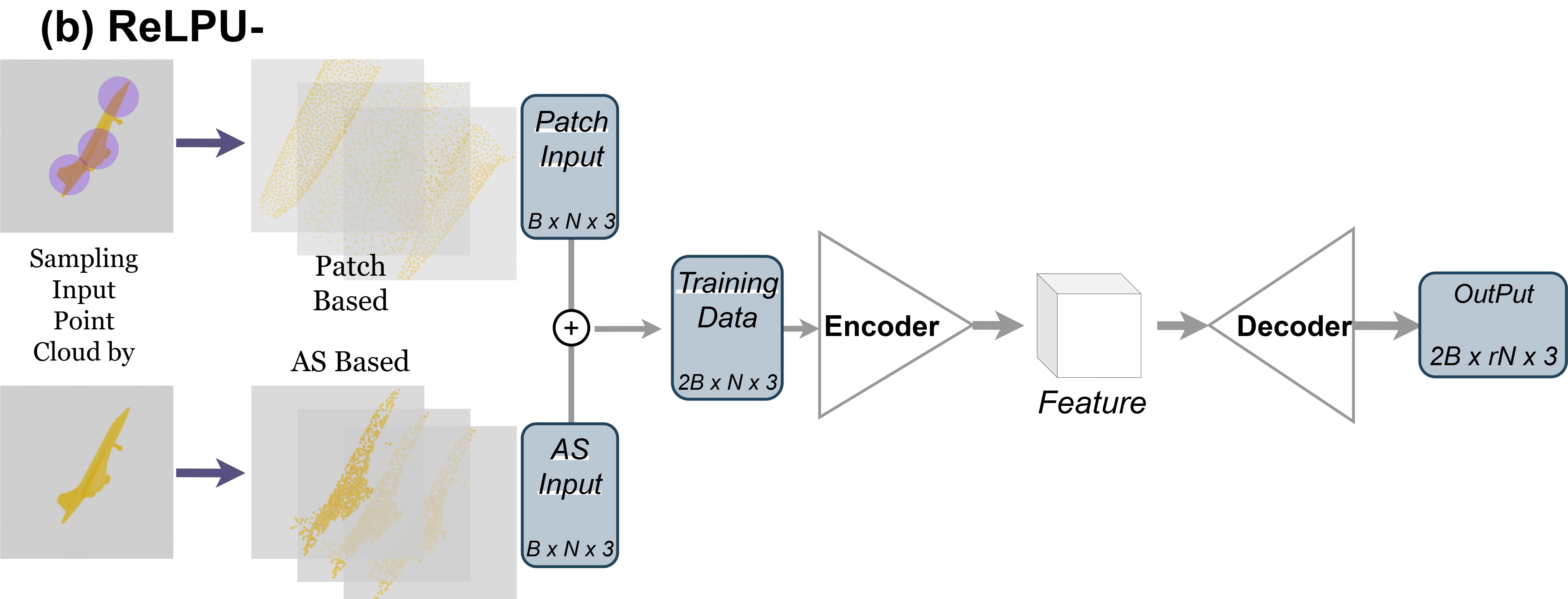}}
		\centerline{\includegraphics[width=\textwidth]{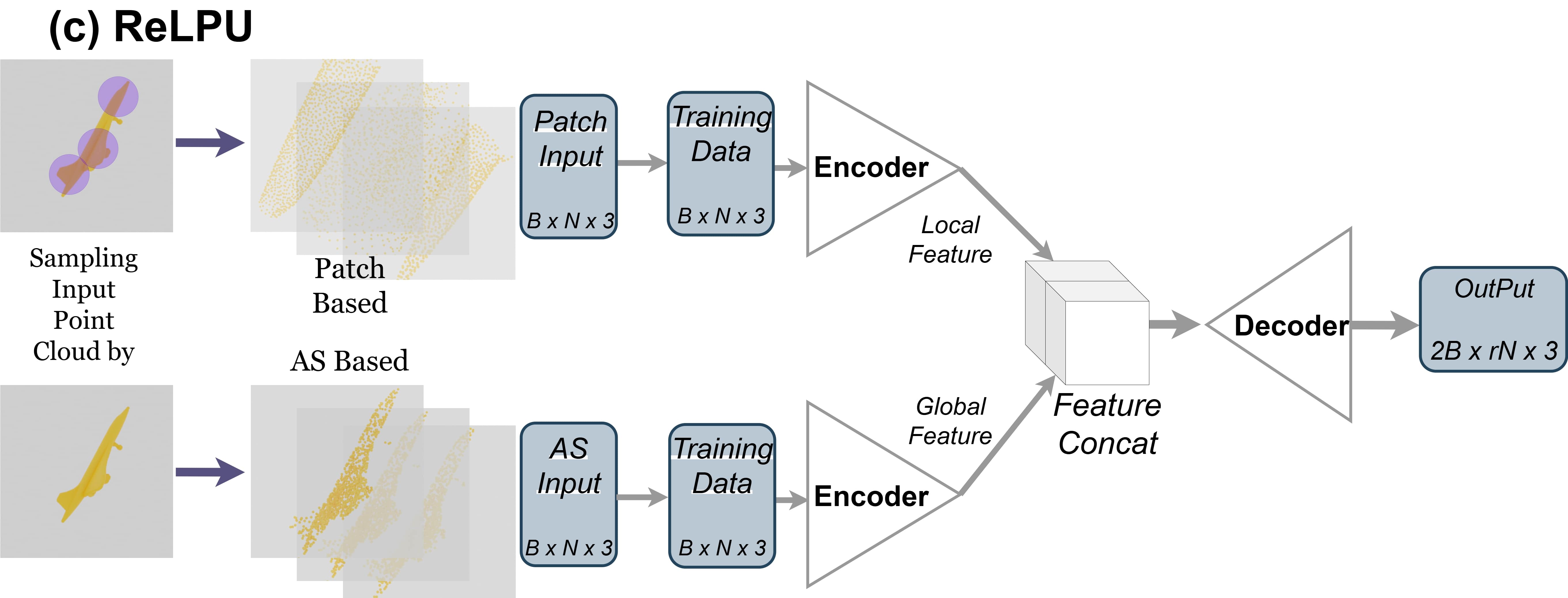}}
        \end{minipage}
    \caption{Illustrates the structure of Traditional Upsampling Network (a) like MPU, PU-GCN, etc. And present schematic diagram of two data input methods as the transcendental local and global feature for ReLPU- (b) and ReLPU (c) models for point cloud upsampling. ReLPU- uses a sequential autoencoder for unified feature extraction, while ReLPU employs two parallel autoencoders to extract local and global features separately, enhancing upsampling performance.}
  \label{fig:1}
\end{figure}

Therefore, we proposed the ReLPU network, adaptable to various encoder-decoder-based point cloud upsampling models. The aim was to address point cloud irregularity and unstructured characteristics by integrating information from both local and global inputs in a parallel decoder. Furthermore, ReLPU enabled easy backbone network replacement in future advanced autoencoder-based point cloud upsampling tasks. That is to say, our proposed framework could be used for continuously updating state-of-the-art point cloud upsampling networks. And there was no need for specific specialized network modules to extract local and global features. We validated our method on the publicly available synthetic datasets PU1K \cite{qian2021pugcn}. In these experiments, our approach was applied to existing networks and demonstrated superior results compared to state-of-the-art methods.

Our main contributions were as follows:

1) We proposed the ReLPU network, a new framework for point cloud upsampling with local and global data input encoders. The parallel encoders extracted local and global features and could be used in various point cloud upsampling networks based on existing autoencoder architectures.

2) Through representation learning, we calculated gradient contribution values to interpret the local and global inputs, effectively explaining the differences and significance of local and global features in point cloud upsampling.

3) Experiments were conducted on the PU1K datasets, establishing baselines for multiple point cloud upsampling models to demonstrate ReLPU's superior performance over previous state-of-the-art methods.

\section{Related Works}

\subsection{Deep Learning in Point Cloud Processing}

Since PointNet \cite{qi2017pointnet, qi2017pointnet++}, early methods that projected 3D point clouds into multi-view 2D images \cite{su2015multi, lawin2017deep} or converted point clouds into voxel or grid discretizations \cite{zhou2018voxelnet, graham2018semantic, guo2020deep} have gradually shifted toward directly processing point cloud data on a point-based basis. This direct processing approach avoids contextual loss and complex steps involved in data conversion. The mainstream point cloud processing methods have evolved from MLP-based modules starting with PointNet \cite{qi2017pointnet} to convolutional neural network-based approaches like PointCNN \cite{li2018pointcnn, shi2019pointrcnn}, and most recently, to Transformer \cite{vaswani2017attention} based architectures such as Point-Transformer \cite{zhao2021point}. As observed, transformer-based structures have already been widely explored in point cloud processing.

\subsection{Similar Approaches in Point Cloud Completion}
Point cloud completion tasks are similar to point cloud upsampling tasks in that they aim to supplement geometric details by extracting global features from partial inputs. With the aid of deep neural networks and extensive 3D datasets, learning-based methods have achieved remarkable performance in shape completion tasks. Similar to point cloud processing, PCN \cite{yuan2018pcn} first used a similar encoder to extract features and output dense and completed point clouds from sparse and incomplete inputs. Agrawal et al. \cite{gurumurthy2019high} employed GAN networks to implement an optimization algorithm for potential noise points. SA-Net \cite{wen2020point} used self-attention mechanisms to effectively leverage local structural details. Lin et al. \cite{lin2024infocd,lin2023hyperbolic} obtained state-of-the-art new results by improving Chamfer distance on multiple benchmark datasets. Zhang et al. \cite{zhang2020detail} proposed a feature aggregation strategy to retain essential details. However, these methods only extract global features from partial inputs, leading to information loss during the encoding process. Later, ASFM-Net \cite{xia2021asfm} used parallel global and local feature matching to reasonably infer the missing geometric details of objects.

\subsection{Current Developments in Point Cloud Upsampling}
Learning-based methods became prevalent and dominated recent state-of-the-art technology. Specifically, PU-Net \cite{yu2018pu} was a pioneering work that introduced CNNs into point cloud upsampling based on the PointNet++ \cite{qi2017pointnet++} backbone. Subsequently, MPU \cite{yifan2019patch} proposed a patch-based upsampling pipeline, allowing flexible upsampling of point cloud patches with rich local details. Additionally, PU-GAN \cite{li2019pugan} applied a generative model to address the issue of high-resolution point cloud generation. PU-GCN \cite{qian2021pugcn} used a graph-based network structure to achieve effective upsampling performance. Dis-PU \cite{li2021point} employed disentangled refinement units to progressively generate high-quality point clouds from coarse point clouds. PU-EVA \cite{luo2021pueva} introduced an edge-vector-based upsampling approach, similar to Grad-PU\cite{he2023gradpu}, to flexibly perform upsampling at different scales. PU-Dense \cite{akhtar2022pu} utilized sparse tensor-based representations to improve the efficiency and scalability of geometry upsampling. PU-Mask \cite{liu2024pu} introduced an implicit virtual mask to guide the generation of detailed and structured point clouds. PU-Refiner \cite{liu2022purefiner} designed a geometry refiner using adversarial learning to enhance the fidelity of upsampled results. Du et al. (PUCRN) \cite{du2022cascaded} and PU-Transformer \cite{qiu2022putransformer} optimized point cloud generation and performance through multi-stage structures and Transformer models.

However, these methods still faced key challenges, especially in sparse or missing regions. The inherent characteristics of point clouds often resulted in models lacking sufficient information to comprehend and describe these regions, leading to overfitting. We advocated for a symmetric, parallel encoder pair that could replace the above models, with global and local feature input on the data side, as shown in Figure \ref{fig:1}, to avoid complex networks and ensure interaction between local and global features, thereby preventing information loss in the feature integration process. To validate this information preservation, we used representation learning to visualize gradient contribution values and output a saliency map \cite{simonyan2013deep}, indicating which points played a critical role in the upsampling process. By using a saliency map to reflect the differences in local and global features between local and global inputs, it indicated that upsampling indeed required prior knowledge of both local and global features.

\section{Methodology}
In this paper, we propose a new Representation Learning of Point Cloud (ReLPU) framework and apply it to point cloud upsampling. Specifically, we construct a parallel network for global and local feature extraction based on the two-point cloud data input methods discussed in \cite{zhang2024rethinking}. In this network, edge attributes are used for detailed geometric patterns, while global attributes aid in a broader understanding of shapes and structures. Finally, we combine geometric features with miscellaneous structural understanding to achieve comprehensive point cloud upsampling. The entire network can be trained in an end-to-end manner. Details are provided in the following sections.

\subsection{Netwotk Architecture}
Based on this fundamental idea, we employ a sequential autoencoder for point cloud upsampling to extract geometric and miscellaneous structural features, resulting in the ReLPU- model (as shown in Figure ~\ref{fig:1}b). We then update ReLPU- by using one specific autoencoder to extract local features and another identical autoencoder to extract other global features, resulting in the ReLPU model (as shown in Figure ~\ref{fig:1}c). This setup enables ReLPU to perform parallel global and local feature extraction, contrasting with the sequential approach in ReLPU-. ReLPU significantly enhances the upsampling performance of ReLPU-.

For the autoencoder, the backbone network can be replaced based on existing point cloud upsampling models, as long as the model adheres to the encoder (feature extraction) to decoder (feature expansion) structure. In this paper, we use MPU \cite{yifan2019patch}, PU-GCN \cite{qian2021pugcn}, Dis-PU \cite{li2021point}, and PUCRN \cite{du2022cascaded} as backbone networks for point cloud feature extraction. This approach aims to validate the effectiveness of our proposed parallel method and to distinguish it from the sequential method. Here, we \textbf{concatenate} the global and local features extracted from the feature extraction modules, or encoders, of different networks.

For local input by patch, assuming the entire Model contains $M$ points, divide it into $K$ patches, so we have $\mathcal{P}_k$ as points of patch input. For global input by Average Segment (AS), once sampling is completed, obtain a point set containing $M$ points $\mathcal{P}_{\text{sampled}}$. We hope to divide it into $K$ subsets, each containing $M/K$ points.

The $k$-th subset is denoted as $\mathcal{P}_k$, where $k=1,2,\cdots,K$. Assuming that $M$ can be divided by $K$, the range of points for each subset is:

\[
\mathcal{P}_k = \{ p_i \mid (k-1) \cdot \frac{M}{K} + 1 \leq i \leq k \cdot \frac{M}{K} \}
\]

After passing through $l$ layers of encoder $E$, we have:

\[
\mathbf{F} = E^{(l)} = \sigma \left( \mathbf{A} E^{(l-1)} \mathbf{W}^{(l)} \right)
\]

where:
\begin{itemize}
    \item $\mathbf{A}$ is the adjacency matrix.
    \item $E^{(0)} = \mathbf{X}$, representing the feature matrix of the input point.
    \item $\sigma$ is the activation function.
\end{itemize}

Additionally, let:
\begin{itemize}
    \item $\mathbf{F}_{\text{global}}$ represent the global feature vector extracted by the autoencoder.
    \item $\mathbf{F}_{\text{local}}$ represent the local feature vector extracted by the autoencoder.
    \item $\mathbf{F}_{\text{concat}}$ represent the concatenated feature vector.
\end{itemize}

To express this concatenation operation, we can define it as follows:

\[
\mathbf{F}_{\text{concat}} = \text{Concat}(\mathbf{F}_{\text{global}}, \mathbf{F}_{\text{local}}) = 
[\mathbf{F}_{\text{global}}, \mathbf{F}_{\text{local}}] \in \mathbb{R}^{m \times 2d}
\]

\begin{figure}[tb]
  \centering
  \includegraphics[height=8cm]{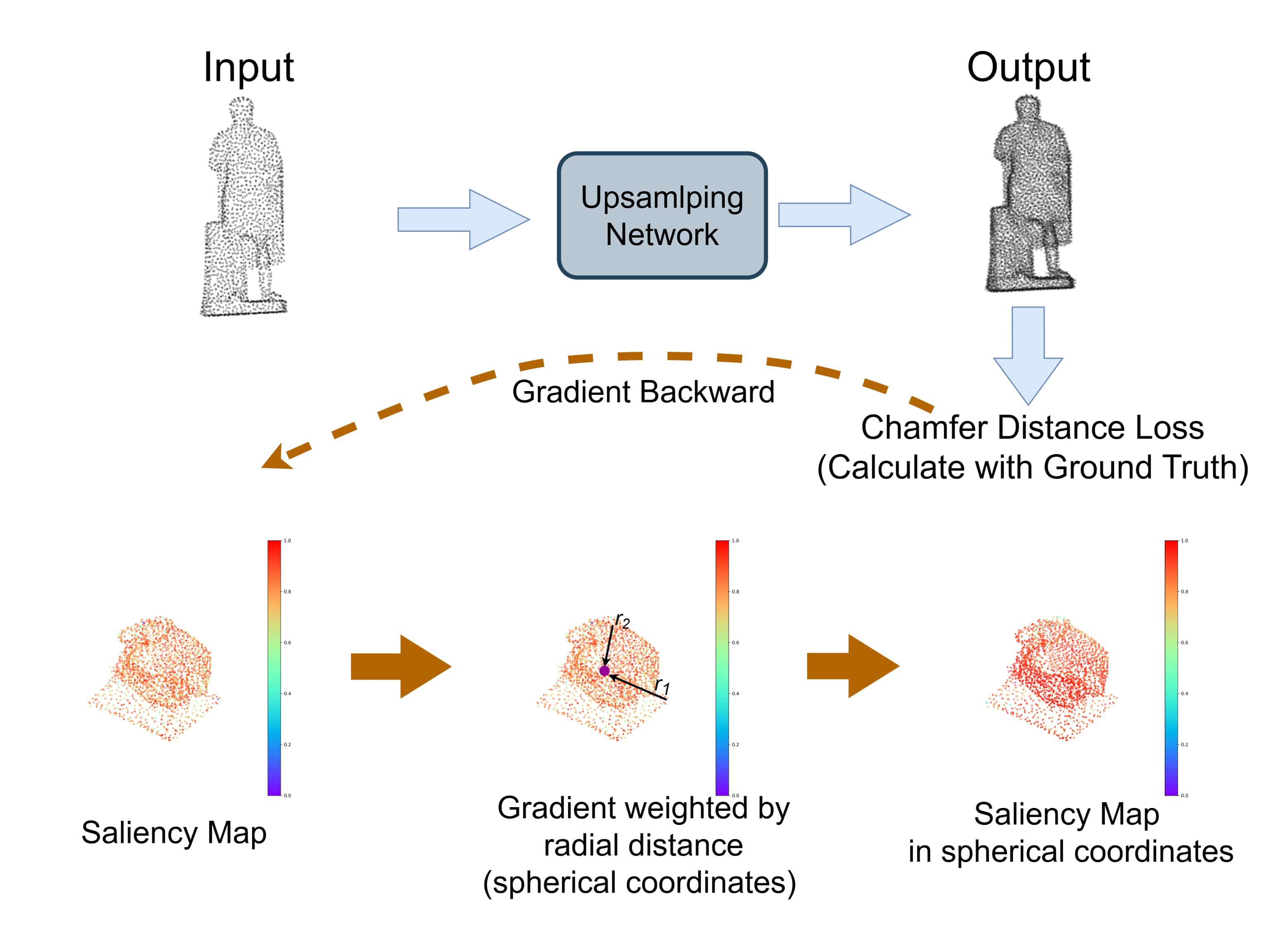}
  \caption{Examples of saliency maps return upsampling loss after multiply the radial distance from the point to the center in spherical coordinates.}
  \label{fig:2}
\end{figure}

\subsection{Saliency Map for Point Cloud} \label{sec:3}
Whether or not one considers the interpretability of deep learning as essential, using saliency maps to calculate gradient contribution values for data analysis and visualization is a key aspect of explainable artificial intelligence (XAI). It provides an intuitive representation of the key factors underlying model results. Saliency maps have been widely applied in point cloud tasks, particularly for classification and detection \cite{zheng2019pointcloud, ziwen2019visualizing, schinagl2022occams}. However, these applications are limited to those tasks. Therefore, to understand the contributions of global and local features to point cloud upsampling and to compare these contributions between sequential and parallel networks, we use saliency maps to reflect the gradients of point clouds during model training, indicating their contribution to upsampling.

Assuming that the loss result that calculate with groundtruth of the upsampling model is defined as $L$, given a point $x_i$ in the point cloud, its contribution can be defined as the difference in loss before and after removing that point:

\[
\Delta L_{x_i} = L(X_{\text{pred}}) - L(X_{\{x_{i_{\text{pred}}}\}})
\]

where:
\begin{itemize}
    \item $X_{\text{pred}}$ represents the upsampling results truth point cloud.
    \item $X_{\{x_{i_{\text{pred}}}\}}$ represents the upsampled point cloud after removing point $x_i$.
\end{itemize}

In other words, the contribution of a point $x_i$ in the point cloud equals the difference between the loss $L(X_{\text{gt}})$ of the complete point cloud of ground truth and the loss $L(X_{\{x_{i_{\text{pred}}}\}})$ of the point cloud after upsampling.

If the contribution value $\Delta L_{x_i}$ is small, we consider point $x_i$ to have a high contribution to the upsampling process, indicating a more accurate resolution after upsampling. Conversely, if $\Delta L_{x_i}$ is large, the contribution of point $x_i$ is considered low. 

To represent this in the saliency map, we assign each point $x_i$ a sensitivity score $s_i$, indicating the contribution level of point $x_i$.

For the \textbf{Saliency Map Expression}, formally, the saliency map can be represented as a function $S_\theta(\cdot)$, where $S_\theta(X)$ takes a point cloud $X$ as input and outputs a vector of length $N$, with each element corresponding to the sensitivity score of each point in the point cloud:

\[
S_\theta(X) = \{s_i \mid i = 1, 2, \ldots, N\}
\]

where the sensitivity score $s_i$ represents the contribution of point $x_i$. If $s_i$ is high (positive), it indicates that the point has a large positive contribution to the upsampling model. If $s_i$ is low or negative, it indicates that the point has a smaller or negative contribution to the upsampling model, as illustrated in Figure \ref{fig:2}, we present the saliency map in input point cloud and the workflow for getting the saliency map.

And then, represent point cloud saliency map as spherical coordinates. Spherical coordinates are defined with the centroid of the point cloud as the reference center, ensuring a consistent measurement of point displacement and contribution. In spherical coordinates, point displacement is considered, where a point is represented as $(r, \psi, \phi)$, with $r$ being the distance from the point to the sphere's center, and $\psi$ and $\phi$ being the two angular coordinates relative to the sphere's center. In this spherical coordinate system, moving a point toward the center by $\delta$ will increase the loss $L$ by an amount:$\ -\frac{\partial L}{\partial r} \delta$

According to the equivalence $\Delta L_{x_i}$, we approximate $\Delta L_{x_i}$ by using a first-order Taylor expansion. So we measure a point's contribution with a real-valued score: the negative gradient of $L$ with respect to $r$:

\[
s_i = -\frac{\partial L}{\partial r_i} r_i^{1+\alpha}.
\]

Thus, in the rescaled coordinate system, we measure the contribution of point $x_i$ simply by using. More precisely, we approximate the contribution of each input point by computing the gradient of the loss with respect to that point, which reflects how much the point affects the upsampling quality. The pseudocode for the above method is showing in Algorithm \ref{alg:1}.

\begin{figure}[t]
\centering
\begin{minipage}[t]{0.85\textwidth}
    \vspace{0pt}
    \footnotesize
    \begin{algorithm}[H]
    \caption{Enhanced Saliency Map Calculation in Spherical Coordinates}\label{alg:1}
    \KwIn{$\mathbf{data}$ (Point cloud), $\mathbf{gt}$ (Ground truth), $\mathbf{out}$ (Model output)} 
    \KwOut{$\mathbf{saliency\_spherical}$ (Saliency map in spherical coordinate)}
    $loss = \text{ChamferDist}(\mathbf{out}, \mathbf{gt})$
    %{\#$N \gets \text{number of objects}$} 

    $loss.\text{backward()}$
    
    $saliency = |\mathbf{data}.grad|$ 

    $centroid = \text{mean}(\mathbf{data})$
    
    $r = \text{norm}(\mathbf{data} - centroid)$
    \tcp*{Calculate radial distance $r$ by matrix norm}
    
    $\mathbf{saliency\_spherical} = -saliency * r$
    \tcp*{Enhance saliency map by radial distance $r$ in spherical coordinate}
    \end{algorithm}
\end{minipage}
\end{figure}

In this case, by incorporating spherical coordinates into saliency maps, this method provides a more interpretable understanding of how global and local features facilitate point cloud upsampling. And points with the same score but different radial distances, and have a smaller gradient indicate that points farther from the center of the circle have stronger initial significance, that suggesting they carry more structural or geometric importance relative to their position. This highlights the effectiveness of the method proposed in this article in capturing changes in structured features.

\subsection{Training Loss Function}
During the training process, our end-to-end network uses Chamfer Distance (CD) as the loss function. CD loss is designed to calculate the average closest point distance between two sets of point clouds, describing the discrepancy between the predicted and actual upsampled points.

For the parallel network ReLPU, it is implemented by fine-tuning ReLPU- to extract both local and global features, with CD loss remaining unchanged.

Given two sets of point clouds $\mathbf{P}$ and $\mathbf{Q}$, representing the predicted and ground truth point clouds, respectively, the CD Loss can be expressed as follows:

Let 
\[
\mathbf{P} = \{p_1, p_2, \ldots, p_M\} \quad \text{and} \quad \mathbf{Q} = \{q_1, q_2, \ldots, q_N\},
\]
where $p_i$ and $q_j$ are points in point clouds $\mathbf{P}$ and $\mathbf{Q}$, respectively. The Chamfer Distance $d_\text{CD}(\mathbf{P}, \mathbf{Q})$ is defined as:

\[
d_\text{CD}(\mathbf{P}, \mathbf{Q}) = \frac{1}{M} \sum_{i=1}^M \min_{j=1,\ldots,N} \|p_i - q_j\|^2 
+ \frac{1}{N} \sum_{j=1}^N \min_{i=1,\ldots,M} \|q_j - p_i\|^2.
\]

This loss function minimizes the distance between each point in one set and its nearest neighbor in the other set, thus effectively capturing the geometric discrepancy between the predicted upsampled point cloud and the ground truth.

\section{Experiments}

\subsection{Settings}

\subsubsection{Dataset}
In our study, we conducted experimental validation on three different datasets: two synthetic dataset \cite{qian2021pugcn, Koch_2019_CVPR} for training and testing, and a real-world object dataset \cite{uy2019revisiting} for visualization. In training and testing part of upsampling, the datasets we used, are the synthetic dataset PU1K dataset \cite{qian2021pugcn} and ABC dataset \cite{Koch_2019_CVPR} for quantitative evaluation. In visualization part of upsampling, the datasets we used, are the synthetic dataset PU1K dataset \cite{qian2021pugcn} and ScanObejctNN \cite{uy2019revisiting} for qualitative evaluation. 

\begin{itemize}
 \item \textbf{PU1K:} This includes 69,000 training samples. PU1K is a new point cloud upsampling dataset introduced in PU-GCN. Overall, PU1K consists of 1,147 3D models, split into 1,020 training samples and 127 testing samples. The training set includes 120 3D models from the PU-GAN dataset \cite{li2019pugan} and 900 different models collected from ShapeNetCore \cite{chang2015shapenet}. The test set contains 27 models from PU-GAN and over 100 models from ShapeNetCore, covering 50 object categories.
\end{itemize}

The original PU1K dataset was modified to fit the patch-based upsampling pipeline, with training data generated through Poisson disk sampling from patches of 3D meshes. Specifically, the original training data comprises 69,000 samples, each containing 256 input points (low resolution) and 1,024 points ($4\times$ high resolution). To match the input model with the number of patches and AS inputs \cite{zhang2024rethinking}, we constrained PU1K to 8,192 points via Poisson disk sampling. Each model was divided into eight different local parts using patches and eight uniform subdivisions with AS to maintain model shape integrity. Similar to previous state-of-the-art point cloud upsampling models, each sample contains 256 input points (low resolution) and 1,024 points ($4\times$ high resolution). Under these conditions, our training data includes 16,320 samples.

\begin{itemize}
\item \textbf{ABC Dataset:} 
The ABC Dataset is a dataset containing one million computer-aided design (CAD) models. This dataset is a collection of explicitly parametrized curves and surfaces for differential quantities, patch segmentation, geometric feature detection, and shape reconstruction.
\end{itemize}

The vast CAD dataset brings diversity, with models covering a wide range of geometric shapes and categories, including architecture, furniture, mechanical parts, and more. The method we proposed will attempt to evaluate the ability of point cloud upsampling networks in geometric feature detection and shape reconstruction on this dataset. This article is based on a subset of ABC 10k in the ABC dataset, which is a CAD model with a quantity of 10000. On this basis, two 4x upsampling experiments were conducted, with the inputs being the complete model and the patch model, respectively. Each sample has low resolution: 512 input points and high resolution: 2048 groundtruth points.

\begin{itemize}
\item \textbf{ScanObjectNN Dataset:} 
This dataset scans real objects, including 15000 objects that are classified into 15 categories with 2902 unique object instances.
\end{itemize}

For ScanObjectiNN, because its data come from real objects, and covers multiple object categories, it helps us comprehensively verify the performance of the network. Nonetheless, the lack of ground truth point clouds in the real scanning dataset, we only conducted qualitative evaluations.

\subsubsection{Evaluation Metrics}
To evaluate our point cloud upsampling network, we used three key metrics: Chamfer Distance (CD), Hausdorff Distance (HD), Point-to-Surface Distance (P2F), and Uniform Metric.

\begin{itemize}
\item CD assesses geometric accuracy through the mean nearest-neighbor distance.
\item HD focuses on the maximum distance mismatch, highlighting extreme cases.
\item P2F examines the proximity of upsampled points to the reference model’s surface, reflecting detail fidelity.
\item Uniform Metric proposed by PU-GAN\cite{li2019pugan}, in order to avoid the local clutter of points and cannot distinguish between different disks containing the same number of points when calculating the overall uniformity of the point set for all objects in the test dataset.
\end{itemize}

Our study inputs all $N$ points in the point cloud. Next, trained model to upscale the seed patches and uniform segments by a scale factor of $r$. The farthest point sampling algorithm \cite{qi2017pointnet} will be used to combine all upsampled patches into a dense output point cloud with $rN$ points. In the $4\times$ upsampling experiments, we performed three low-to-high-resolution sampling conversions. Each test sample in PU1k has a low-resolution point cloud with 512, and 2,048 points and a high-resolution point cloud with 2,048, and 8,196 points. For the test sample in ABC has a low-resolution point cloud with 512 points and upsampling to high-resolution point cloud with 2,048.

%\begin{figure}[tb]
%  \centering
%  \includegraphics[width=\textwidth, height=15cm]{figure3.jpg}
%  \caption{Visualizes saliency maps comparing ReLPU and original models (MPU, Dis-PU, PU-GCN, PUCRN) by patch (local) inputs and AS (global) inputs. Points are color-coded based on normalized saliency score ranks in spherical coordinates, where higher values represent higher saliency.  }
%  \label{fig:3}
%\end{figure}

\begin{figure}[tb]
  \centering
	\begin{minipage}{0.48\linewidth}
		\centerline{\includegraphics[width=\textwidth]{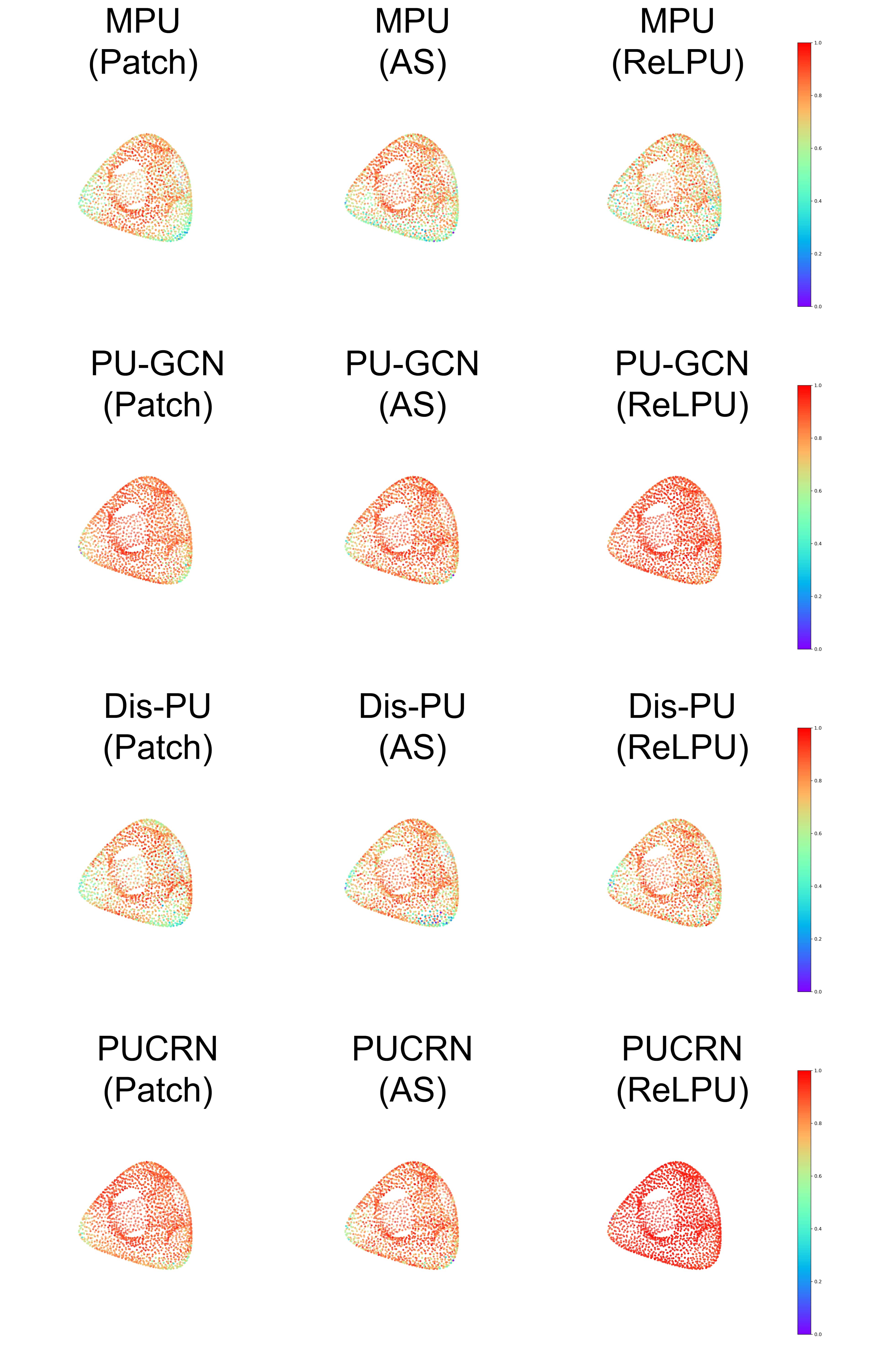}}
            \centerline{Model: genus3}
	\end{minipage}
	\begin{minipage}{0.48\linewidth}
		\centerline{\includegraphics[width=\textwidth]{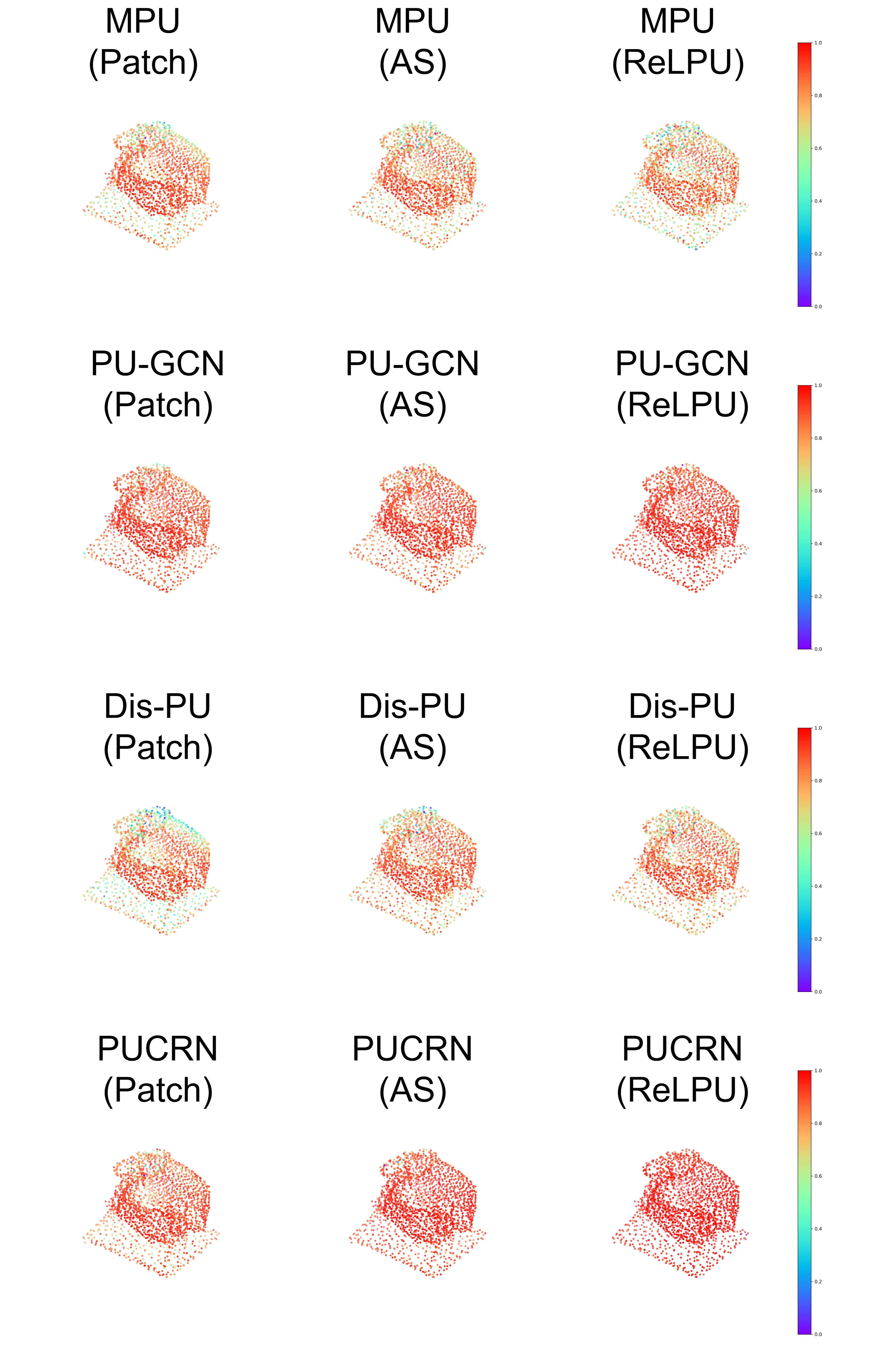}}
            \centerline{Model: statue\_rome\_aligned}
        \end{minipage}
  \caption{Visualizes saliency maps comparing ReLPU and original models (MPU, Dis-PU, PU-GCN, PUCRN) by patch (local) inputs and AS (global) inputs. Points are color-coded based on normalized saliency score ranks in spherical coordinates, where higher values represent higher saliency.
  }
  \label{fig:3.1}
\end{figure}

\begin{figure}[tb]
  \centering
  \includegraphics[width=0.9\textwidth]{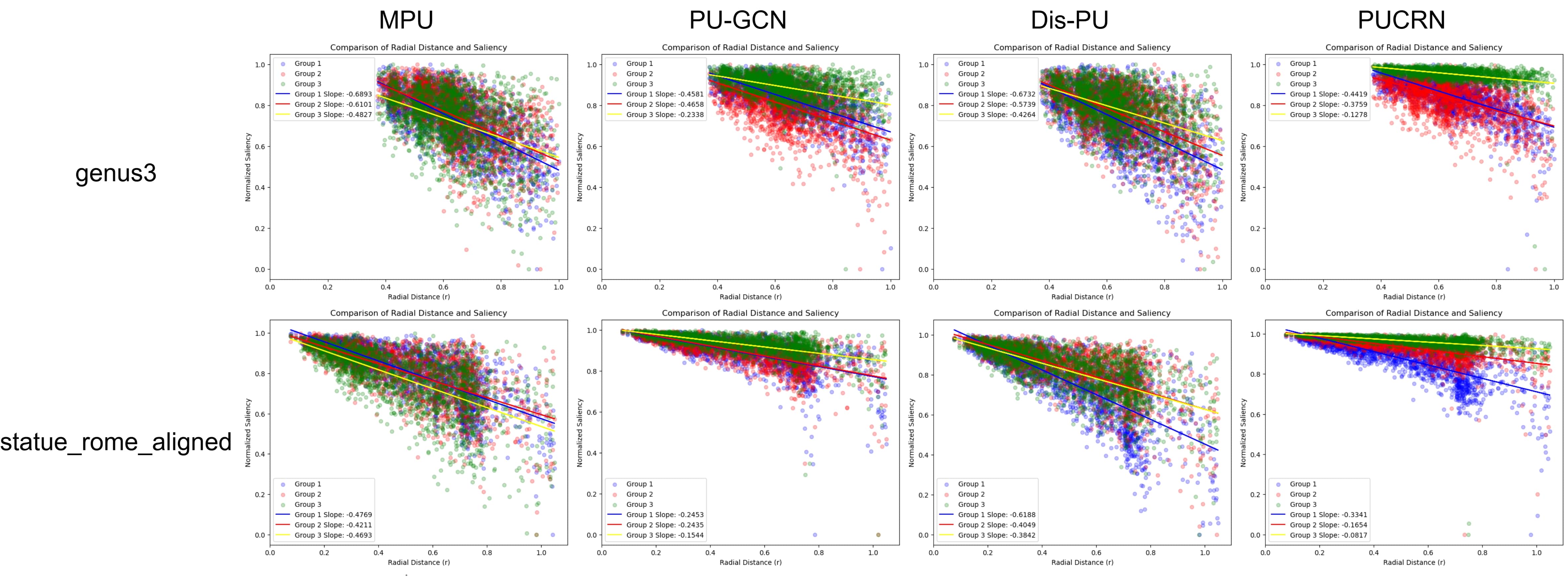}
  \caption{Simple Linear regression of saliency maps comparing original models (MPU, Dis-PU, PU-GCN, PUCRN) by patch (local) inputs: Group 1; and AS (global) inputs: Group 2; and ReLPU: Group 3. Where $S_\theta(X) = - x_i * r$, the slope is represented as $x_i$. When $x_i$ tends towards 0, under the same $s_i$ with farther radial distance $r$ will have larger saliency maps score in $x_i$.} 
  \label{fig:3.2}
\end{figure}

\subsubsection{Comparison Methods}
To verify the effectiveness of our proposed point cloud upsampling method, we compared it with several advanced algorithms: MPU \cite{yifan2019patch}, PU-GCN \cite{qian2021pugcn}, Dis-PU \cite{li2021point}, and PUCRN \cite{du2022cascaded}. We also compared these algorithms with the ReLPU network using the same training pipeline.

For a fair and objective comparison, we obtained the open-source codes of these methods and trained and tested them on our computing equipment and training framework with identical parameter settings to establish the baseline.

\subsubsection{Implementation Details}
Our network was developed using the PyTorch framework and ran on an Ubuntu 22.04 system. We used an NVIDIA L20 GPU with 48GB of graphics memory and an Intel(R) Xeon(R) Platinum 8457C host with 100GB of RAM. The network was trained over 100 epochs with a batch size of 32. The initial learning rate was set to 0.0005, with a decay rate of 0.05.

\begin{figure}[tb]
    \centering
    \begin{minipage}{0.9\linewidth}
		\centerline{\includegraphics[width=\textwidth]{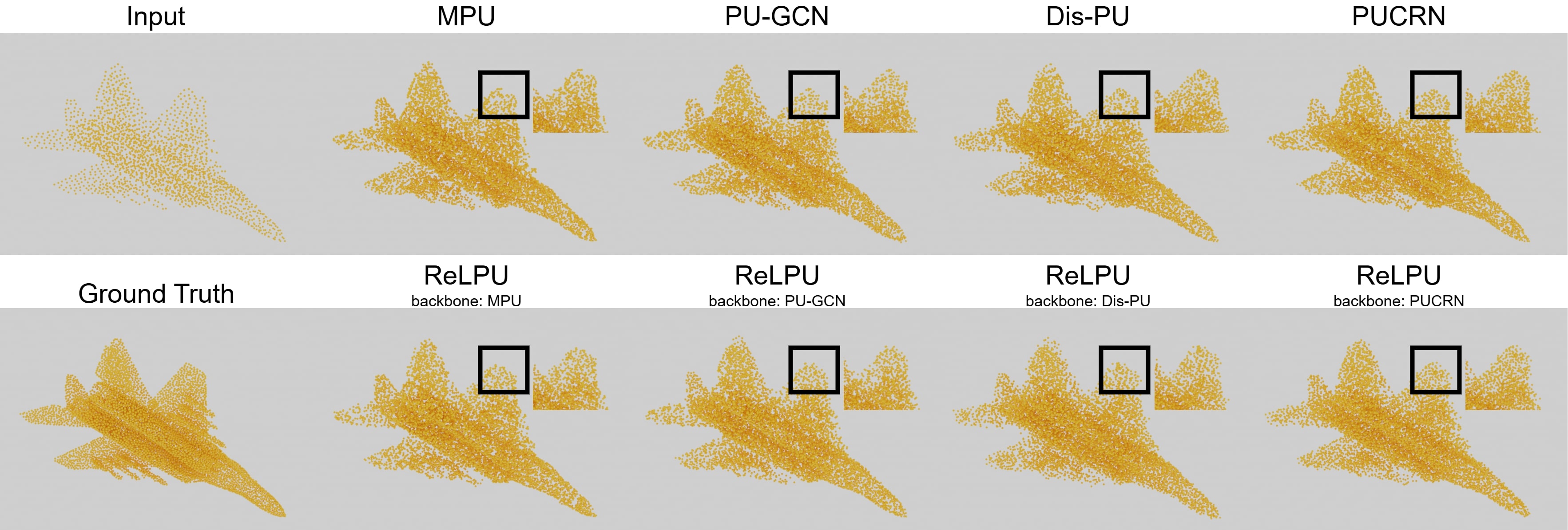}}
		\centerline{\includegraphics[width=\textwidth]{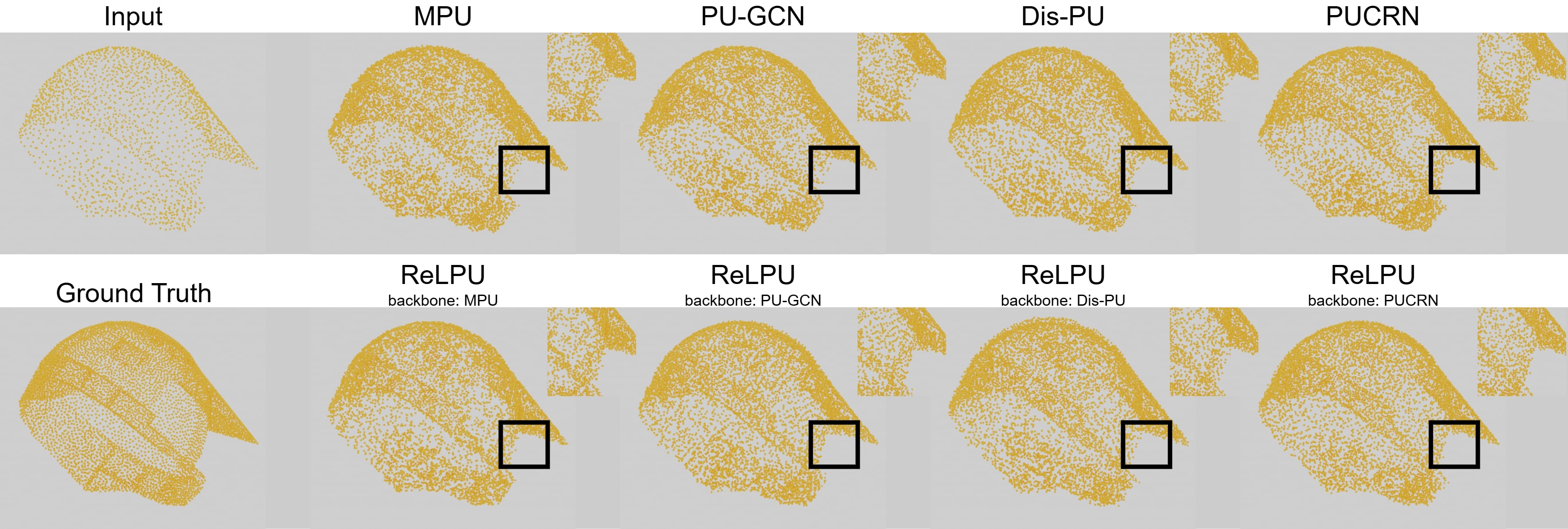}}
        \end{minipage}
    \caption{Visualize the result of state-of-the-art methods (MPU, PU-GCN, Dis PU and PUCRN) with ReLPU in ($4\times$) upsampling on PU1K test data by using 2048 input points.}
  \label{fig:4}
\end{figure}

%\begin{figure}[tb]
%  \centering
%  \includegraphics[width=\textwidth, height=14.5cm]{figure4.jpg}
%  \caption{Comparisons to state-of-the-art methods (MPU, PU-GCN, Dis PU and %PUCRN) with ReLPU in ($4\times$) upsampling synthetic point cloud data using 2048 input points.  }
%  \label{fig:4}
%\end{figure}

\subsection{Upsampling Results}
In accuracy analysis, as mentioned above, we extensively evaluated the generalization performance of our network on the PU1K synthetic dataset, focusing on the differences between the original backbone network and the ReLPU pipeline incorporating both local and global features. The quantitative results for the PU1K dataset \cite{qian2021pugcn} in Table \ref{tab:1.1} show that our method performs consistently across different input scales (512, 2,048 points) and outperforms the original network in CD, HD, and P2F metrics. Meanwhile, we also showed in Table \ref{tab:1.3} that our method performs under 2048 input scales. And it demonstrates both Uniformity and Complexity.

This result demonstrates the effectiveness of ReLPU, showcasing its superior capability in minimizing nearest-neighbor distance differences and reducing maximum point-to-point distance errors. Additionally, our method achieves low point-to-surface differences, which is crucial for maintaining topological consistency and mesh reconstruction from point clouds. However, compared to PUCRN [25], applying ReLPU to PUCRN yields slight differences in CD, HD, and P2F across the three input scales. This is related to PUCRN's sequential three-stage refinement, which minimizes local and global discrepancies, as detailed in Section \ref{sec:4}. But, ReLPU has improved the performance of the original model in Uniformity. However, in terms of complexity, ReLPU considers separate encoders for local and global inputs, which increases the overall parameters of the model.

\begin{table}[]
\centering
\scalebox{0.75}{
\begin{tabular}{lllllll}
\hline
\multirow{2}{*}{Methods} & \multicolumn{3}{l}{512 input points}               & \multicolumn{3}{l}{2048 input points}             \\ \cline{2-7} 
                         & CD$\downarrow$             & HD$\downarrow$               & P2F$\downarrow$              & CD$\downarrow$              & HD$\downarrow$               & P2F$\downarrow$             \\ \hline
MPU \cite{yifan2019patch}             & 4.973          & 48.403          & 29.357          & 1.217          & 29.283          & 15.823         \\
PU-GCN \cite{qian2021pugcn}          & 4.142          & 43.529          & 21.035          & 0.987          & 23.812          & 10.291         \\
Dis-PU \cite{li2021point}          & 4.120          & 41.665          & 20.578          & 1.025          & 18.238          & 8.189          \\
PUCRN \cite{du2022cascaded}           & \textbf{3.605$\downarrow$} & 35.778          & 15.246          & 0.918          & 13.870          & 6.633          \\
ReLPU (MPU)              & 4.528          & 45.769          & 25.894          & 1.128          & 25.364          & 13.086         \\
ReLPU (PU-GCN)           & 4.002          & 39.527          & 19.529          & 0.953          & 20.533          & 8.317          \\
ReLPU (Dis-PU)           & 3.934          & 39.471          & 19.170          & 0.985          & 17.298          & 7.744          \\
ReLPU (PUCRN)            & 3.625          & \textbf{34.873$\downarrow$} & \textbf{14.802$\downarrow$} & \textbf{0.910$\downarrow$} & \textbf{13.117$\downarrow$} & \textbf{5.682$\downarrow$} \\ \hline
\end{tabular}
}
\caption{Performance comparison of different methods with ReLPU in PU1K across input scales. Among them, the unit is $10 ^ {-3}$.}
\label{tab:1.1}
\end{table}

In addition, we extensively evaluated the generalization performance of our network on the ABC 10k large-scale CAD dataset \cite{Koch_2019_CVPR}, focusing on the differences between the original patch based backbone network and the ReLPU pipeline that combines local and global features. The quantitative results of the ABC 10k dataset in Table \ref{tab:1.2} indicate that, with 512 input points, the ReLPU proposed in this paper outperforms the original network on CD and HD.
However, overall, the values of CD and HD are much higher than those in the PU1K dataset. There are multiple reasons for this. It may be that point cloud upsampling is not suitable for CAD datasets, and geometric features cannot be accurately identified. It may also be due to the small size of the CAD model, which indicates that the point cloud values are too small, resulting in large errors after upsampling and inaccurate results. More details will discuss in Section \ref{sec:5}.

% \usepackage{multirow}
\begin{table}[]
\centering
\scalebox{0.68}{
\begin{tabular}{llllllllll}
\hline
\multirow{2}{*}{Methods} & \multicolumn{3}{l}{2048 input points}             & \multicolumn{5}{l}{Uniformity for Different $p\,(10^{-3})$ $\downarrow$} & Params. \\ \cline{2-9}
                         & CD$\downarrow$ & HD$\downarrow$ & P2F$\downarrow$ & 0.40$\%$               & 0.60$\%$               & 0.80$\%$               & 1.00$\%$               & 1.20$\%$               & (Kb)    \\ \hline
MPU \cite{yifan2019patch}             & 1.217          & 29.283         & 15.823          & 7.51                 & 7.41                 & 8.35                 & 9.62                 & 11.13                & 76      \\
PU-GCN \cite{qian2021pugcn}          & 0.987          & 23.812         & 10.291          & 3.10                  & 3.55                 & 4.59                 & 4.39                  & 4.93                 & 76      \\
Dis-PU \cite{li2021point}          & 1.025          & 18.238         & 8.189           & 2.49                 & 2.53                 & 2.89                 & 3.57                 & 4.21                 & 1047    \\
PUCRN \cite{du2022cascaded}           & 0.918          & 13.870          & 6.633           & 2.48                 & 2.44                 & 2.83                  & 3.45                 & 4.21                  & 847     \\
ReLPU (MPU)              & 1.128          & 25.364         & 13.086          & 6.95                 & 7.12                  & 8.12                 & 9.05                 & 10.74                & 117     \\
ReLPU (PU-GCN)           & 0.953          & 20.533         & 8.317           & 2.88                 & 3.26                 & 4.15                 & 4.55                  & 5.10                  & 120     \\
ReLPU (Dis-PU)           & 0.985          & 17.298         & 7.744           & 2.45                 & 2.49                 & 2.85                 & 3.42                 & 4.01                 & 1628    \\
ReLPU (PUCRN)            & 0.910           & 13.117         & 5.682           & 2.41                 & 2.39                 & 2.75                 & 3.37                 & 3.95                 & 1411    \\ \hline
\end{tabular}
}
\caption{Overall performance under 2048 input points of different methods with ReLPU in PU1k. Including Uniformity and Complexity.}
\label{tab:1.3}
\end{table}

\begin{table}[]
\centering
\scalebox{0.75}{
\begin{tabular}{lll}
\hline
\multirow{2}{*}{Methods} & \multicolumn{2}{l}{512 input points} \\ \cline{2-3} 
                         & CD$\downarrow$               & HD$\downarrow$                \\ \hline
MPU                      & 98.188           & 160.709           \\
PU-GCN                   & 36.998           & 86.791            \\
Dis-PU                   & 30.724           & 59.480            \\
PUCRN                    & 25.873           & 61.993            \\
ReLPU (MPU)              & 80.748           & 110.203           \\
ReLPU (PU-GCN)           & 39.765           & 79.510            \\
ReLPU (Dis-PU)           & 28.351           & 55.942            \\
ReLPU (PUCRN)            & \textbf{20.585$\downarrow$}   & \textbf{42.122$\downarrow$}  \\ \hline
\end{tabular}
}
\caption{Performance comparison of different methods with ReLPU in ABC 10k. Among them, the unit is $10 ^ {-3}$.}
\label{tab:1.2}
\end{table}

\subsection{Noise Robustness Test}

\begin{figure}[tb]
  \centering
  \includegraphics[width=\textwidth]{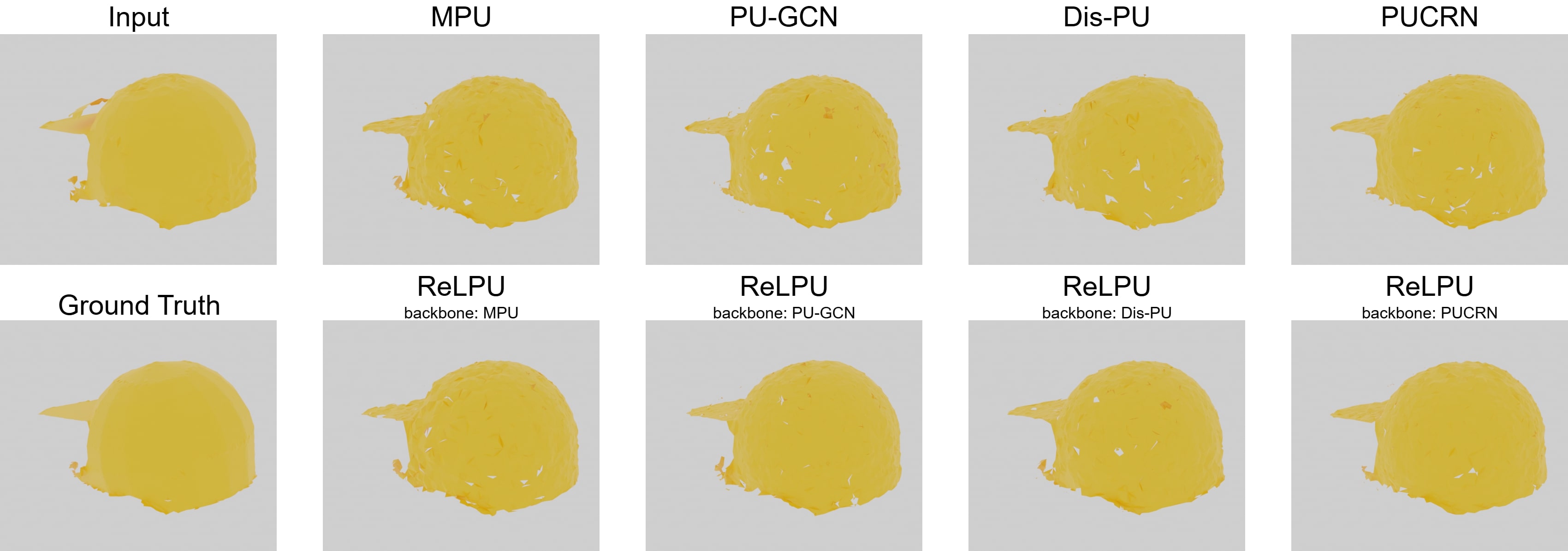}
  \caption{Reconstruct the mesh from PU1K after upsamling by using BallPivoting \cite{bernardini1999ball} algorithm. Our method can be less holes.
  }
  \label{fig:5}
\end{figure}

\begin{figure}[tb]
    \centering
    \begin{minipage}{0.75\linewidth}
		\centerline{\includegraphics[width=\textwidth]{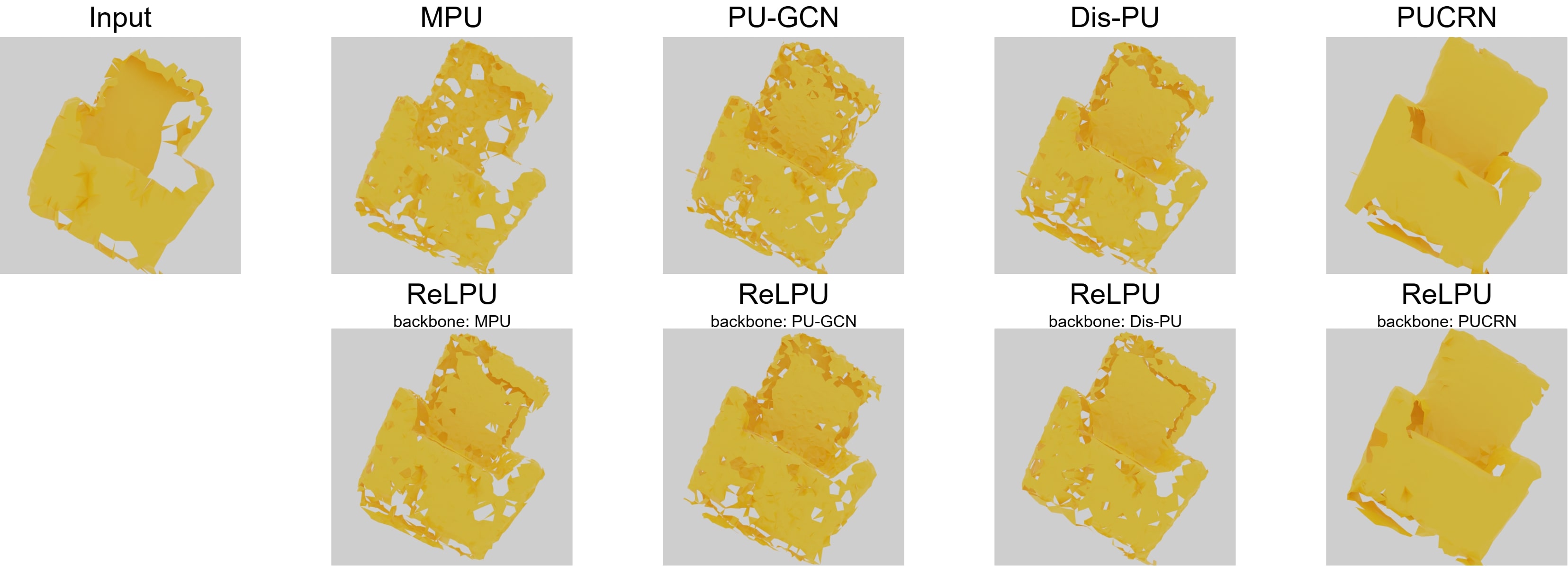}}
		\centerline{\includegraphics[width=\textwidth]{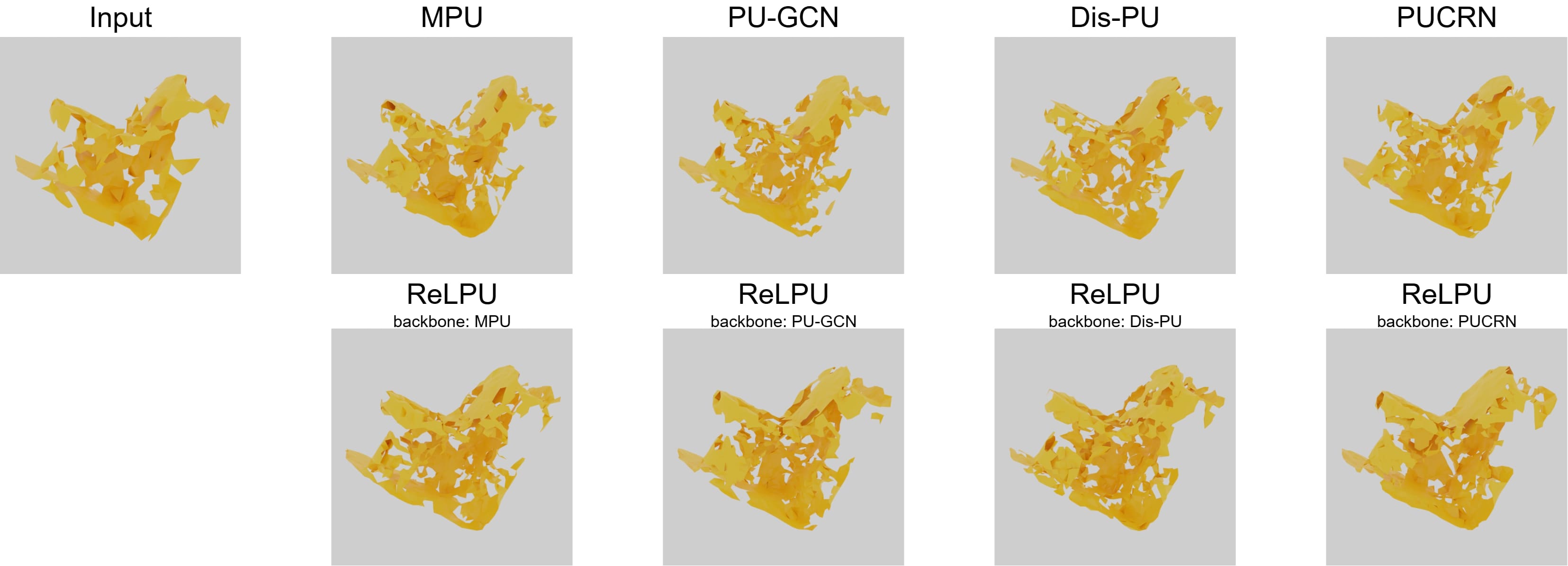}}
        \end{minipage}
    \caption{Reconstruct the mesh from ScanObjectNN after upsamling by using BallPivoting \cite{bernardini1999ball} algorithm. Our method can be smoother and more complete.}
  \label{fig:6}
\end{figure}

It is necessary to verify our model's robustness to noise. Specifically, we tested the pre-trained model by adding random noise to sparse input data, with Gaussian noise satisfying a normal distribution $\mathcal{N}(0,1)$ and scaled by a coefficient. We experimented at two noise levels: $\beta=1\%$ and $\beta=2\%$, in 512 and 2048 points input of PU1K dataset. Table \ref{tab:2.1} and \ref{tab:2.2} quantitatively compares the results of our model and state-of-the-art methods under different noise levels. In most test cases with noise, our proposed ReLPU outperformed the base models MPU \cite{yifan2019patch}, PU-GCN \cite{qian2021pugcn}, Dis-PU \cite{li2021point}, and PUCRN \cite{du2022cascaded} showed robustness in CD and HD metrics.

\begin{table*}[]
\centering
\scalebox{0.7}{
\begin{tabular}{lllll}
\hline
\multirow{2}{*}{Methods} & \multicolumn{2}{l}{\textbf{$\beta=1\%$}} & \multicolumn{2}{l}{\textbf{$\beta=2\%$}} \\ \cline{2-5} 
                         & CD$\downarrow$     & HD$\downarrow$      & CD$\downarrow$     & HD$\downarrow$      \\ \hline
MPU \cite{yifan2019patch}             & 5.632              & 51.334              & 8.054              & 71.868              \\
PU-GCN \cite{qian2021pugcn}          & 4.588              & 44.628              & 6.928              & 51.322              \\
Dis-PU \cite{li2021point}          & 4.272              & 42.107              & 5.938              & 50.528              \\
PUCRN \cite{du2022cascaded}           & \textbf{3.610$\downarrow$}     & 37.086              & 5.514     & 48.212              \\
ReLPU (MPU)              & 4.959              & 49.211              & 7.934              & 66.403              \\
ReLPU (PU-GCN)           & 4.127              & 43.692              & 6.882              & 55.926              \\
ReLPU (Dis-PU)           & 4.080              & 41.824              & 6.013              & 48.098              \\
ReLPU (PUCRN)            & 3.617              & \textbf{36.253$\downarrow$}     & \textbf{5.164$\downarrow$}     & \textbf{47.129$\downarrow$}     \\ \hline
\end{tabular}
}
\caption{Robustness of models tested on 512 points input of PU1K with random noise, where noise follows a normal distribution $\mathcal{N}(0,1)$, and $\beta$ is the noise level. Among them, the unit is $10 ^ {-3}$.}
\label{tab:2.1}
\end{table*}

\begin{table*}[]
\centering
\scalebox{0.7}{
\begin{tabular}{lllll}
\hline
\multirow{2}{*}{Methods} & \multicolumn{2}{l}{\textbf{$\beta=1\%$}} & \multicolumn{2}{l}{\textbf{$\beta=2\%$}} \\ \cline{2-5} 
                         & CD$\downarrow$     & HD$\downarrow$      & CD$\downarrow$     & HD$\downarrow$      \\ \hline
MPU \cite{yifan2019patch}             & 1.634              & 34.030              & 2.276              & 47.302              \\
PU-GCN \cite{qian2021pugcn}          & 1.329              & 28.652              & 1.807              & 32.950              \\
Dis-PU \cite{li2021point}          & 1.305              & 26.174              & 1.792              & 34.026              \\
PUCRN \cite{du2022cascaded}           & 1.276              & 17.502              & 1.726              & 26.578              \\
ReLPU (MPU)              & 1.535              & 33.756              & 2.097              & 44.258              \\
ReLPU (PU-GCN)           & 1.268              & 27.221              & 1.865              & 33.183              \\
ReLPU (Dis-PU)           & 1.231              & 25.553              & 1.723              & 32.733              \\
ReLPU (PUCRN)            & \textbf{1.180$\downarrow$}     & \textbf{16.948$\downarrow$}     & \textbf{1.719$\downarrow$}     & \textbf{25.422$\downarrow$}     \\ \hline
\end{tabular}
}
\caption{Robustness of models tested on 2048 points input of PU1K with random noise, where noise follows a normal distribution $\mathcal{N}(0,1)$, and $\beta$ is the noise level. Among them, the unit is $10 ^ {-3}$.}
\label{tab:2.2}
\end{table*}

\subsection{Ablation Study} \label{sec:4}
To verify the effectiveness of ReLPU compared to the original models, we visualized saliency maps in ReLPU across four models. To better understand our saliency maps, several maps are visualized in Figure \ref{fig:3.1}, where we color-code points based on their saliency score ranks, the saliency maps score has been normalized to $(0, 1)$, with color close to red indicating higher saliency scores. We can see that the four models MPU \cite{yifan2019patch}, PU-GCN \cite{qian2021pugcn}, Dis-PU \cite{li2021point}, and PUCRN \cite{du2022cascaded} have higher eigenvalues under ReLPU method. This means that regardless of the radial distance, each point is repeatedly considered during ReLPU. However, traditional sampling methods, such as patch based methods, score lower at distant edge points, indicating insufficient contribution.

In order to more intuitively represent the relationship in Section \ref{sec:3}, between the $s_i$ of the saliency maps score in spherical coordinates, i.e. $S_\theta(X)$, and the original saliency maps score $x_i$ and the radial distance $r$ in spherical coordinates. We performed a simple linear regression on $r$ and $s_i$. In this case, the slope is represented as $x_i$. When $x_i$ becomes smaller and tends towards 0, under the same $s_i$, the contribution of the larger radial distance $r$ saliency maps score away from the center is larger. As shown in Figure \ref{fig:3.2}, in the simple linear review mentioned above, the higher the radial distance, the higher the ReLPU score proposed in this paper on the spherical saliency maps. Indicating the ability to reuse edge points that are far from the center. Meanwhile, we can see that traditional patch based methods do indeed ignore the contribution of edge points in saliency maps.

We compared ReLPU- and ReLPU to demonstrate the effectiveness of parallel feature processing over sequential processing. Results are shown in the Table \ref{tab:3}. It can be seen that ReLPU based on parallel local and global inputs performs better than ReLPU based on sequential networks, which fully demonstrates the importance of parallel local and global feature fusion.

\begin{table}[]
\centering
\scalebox{0.9}{
\begin{tabular}{llllllll}
\hline
                  & Methods         & \multicolumn{3}{l}{ReLPU-} & \multicolumn{3}{l}{ReLPU}                         \\ \hline
Backbone          &                 & CD$\downarrow$     & HD$\downarrow$      & P2F$\downarrow$     & CD$\downarrow$             & HD$\downarrow$              & P2F$\downarrow$            \\ \hline
\multicolumn{2}{l}{MPU \cite{yifan2019patch}}    & 1.185  & 27.830  & 15.072  & 1.128          & 25.364          & 13.086         \\
\multicolumn{2}{l}{PU-GCN \cite{qian2021pugcn}} & 0.974  & 22.516  & 9.851   & 0.953          & 20.533          & 8.317          \\
\multicolumn{2}{l}{Dis-PU \cite{li2021point}} & 0.979  & 18.001  & 7.969   & 0.985          & 17.298          & 7.744          \\
\multicolumn{2}{l}{PUCRN \cite{du2022cascaded}}  & 0.912  & 13.509  & 5.863   & \textbf{0.910$\downarrow$} & \textbf{13.117$\downarrow$} & \textbf{5.682$\downarrow$} \\ \hline
\end{tabular}
}
\caption{Performance comparison of ReLPU- and ReLPU methods in 2048 input to 8192 scales. Among them, the unit is $10 ^ {-3}$.}
\label{tab:3}
\end{table}

\subsection{Visualization}
Figure \ref{fig:4} provides a qualitative analysis on the PU1K dataset \cite{qian2021pugcn}, demonstrating the effectiveness of our method in preserving overall shape, contour boundaries, and topological features, particularly in the square frame areas. ReLPU can ensure edge integrity and smooth contour at the wing and corner respectively. This is critical for applications that require surface reconstruction with high fidelity and topological accuracy. 

In the Figure \ref{fig:5} and \ref{fig:6}, 2,048 sparse points from PU1K \cite{qian2021pugcn} and ScanObjectNN \cite{uy2019revisiting} are used as input, generating a dense point cloud with 8,192 points at a $4\times$ upsampling ratio. It shows the corresponding 3D mesh reconstructed using the Ball Pivoting algorithm \cite{bernardini1999ball}. From the reconstruction results, although MPU \cite{yifan2019patch}, PU-GCN \cite{qian2021pugcn}, and Dis-PU \cite{li2021point} achieved state-of-the-art (SOTA) performance, they tend to overfit and produce holes when processing sparse boundaries of objects. In the Figure \ref{fig:5},  It can be seen that the ReLPU method can effectively reduce holes and edge noise. The reconstructed helmet is more complete. In the Figure \ref{fig:6}, the ReLPU method shows fewer holes and complete cuts on the sofa under real objects based on ScanObjectiNN \cite{uy2019revisiting}. 

The results indicate that our method significantly improves overfitting in handling sparse and boundary regions. Our method not only enhances the uniformity and detail of point clouds but also substantially improves the accuracy of reconstructing holes and sparse regions, demonstrating the utility of our comprehensive approach in maintaining topological consistency while processing both local and global information.

\section{Limitation and Discussion}\label{sec:5}

\subsection{Limitation}
Although the ReLPU framework performs well in point cloud upsampling tasks, we are also aware of its potential limitations that may affect its applicability and performance in certain scenarios.

\subsubsection{Dependence on input data quality}
\begin{figure}[tb]
    \centering
    \includegraphics[width=0.8\textwidth]{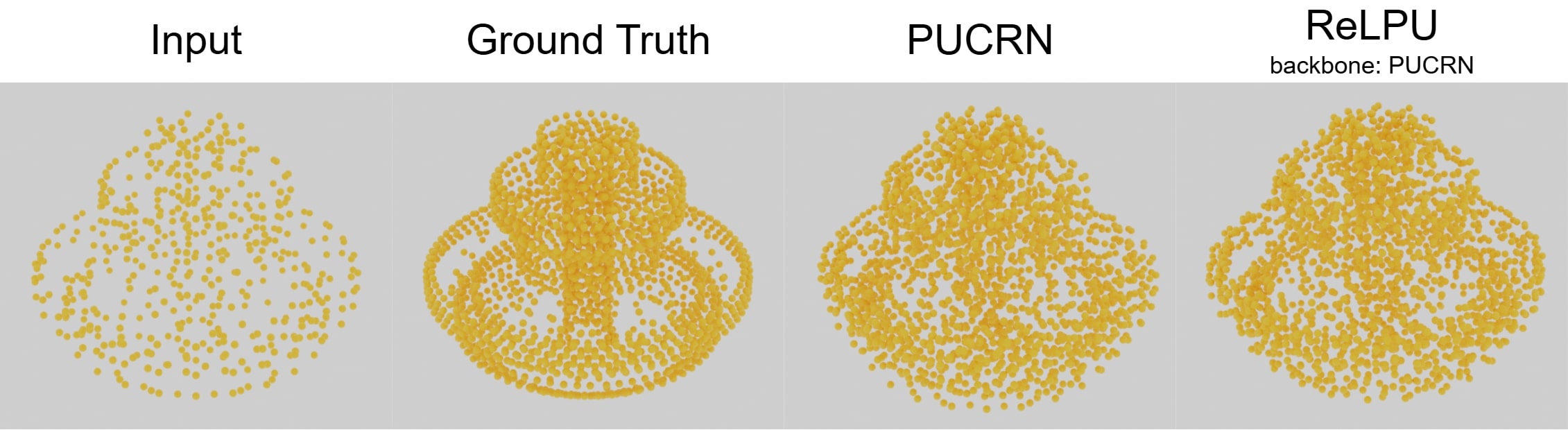}
    \caption{Visualize the result of state-of-the-art methods: PUCRN with ReLPU in ($4\times$) upsampling on ABC test data by using 512 input points, perform unsatisfied result in geometric shapes.}
  \label{fig:7}
\end{figure}

Although ReLPU can effectively handle sparsity and noise issues, its performance still depends on the quality of the input point cloud. In extreme cases, such as the ABC dataset\cite{Koch_2019_CVPR}, where the input point cloud is very regular and the overall volume of the point cloud model is extremely small, the performance of the model may be affected to some extent. Especially for simple but very regular point clouds, the fusion of global and local features may not be ideal, as shown in the Figure \ref{fig:7}.

\subsubsection{The Challenge of Generalization Ability}
Although we have validated the superior performance of ReLPU on the PU1K dataset \cite{qian2021pugcn}, in practical applications, the distribution and characteristics of point cloud data may vary depending on different collection devices and environments. Point cloud upsampling lacks upsampling data from real objects and ground truth \cite{kwon2023deep,zhang2022pointreview}. Although ScanObjectiNN \cite{uy2019revisiting} can be transformed into a point cloud upsampling dataset, each object is incomplete and more suitable for point cloud completion. There may be potential cross domain issues with limited datasets in the field of point cloud upsampling. Therefore, the generalization ability of ReLPU across datasets or domains still needs further validation.

\subsection{Discussion}
With the rapid advancement of deep learning, numerous novel architectures have emerged, such as Transformer \cite{vaswani2017attention}, Mamba\cite{gu2023mamba}, RWKV\cite{peng2023rwkv}, and xLSTM\cite{beck2024xlstm}. These methods have also inspired the development of hybrid approaches, combining their strengths with Graph Neural Networks (GNNs), resulting in architectures like Graph-Transformer\cite{min2022transformer}, Graph-RWKV\cite{he2024pointrwkv} and Graph-Mamba\cite{wang2024graph}. Moreover, there have been attempts to apply these new architectures to point cloud upsampling, exemplified by MBPU\cite{song2024mbpu}, which integrates Mamba into this domain.

Our approach ensures that future innovations in backbone architectures, such as those based on Transformer, Mamba, or other emerging models, can be seamlessly integrated into ReLPU. This adaptability not only future-proofs the framework but also ensures its relevance as point cloud processing techniques continue to evolve. The saliency map reflects the shape of the input data, as well as the local and global shapes, and can indeed represent local and global features, thus achieving the extraction of both types of features in conjunction with the encoder at the input end. Through comprehensive experiments and analyses, we demonstrated the effectiveness of ReLPU in achieving superior geometric fidelity and robustness compared to state-of-the-art methods, making it a promising solution for point cloud upsampling in the era of rapidly evolving deep learning architectures.

In the application field, we also expect that parallel local and global input methods can be widely applied in geodesic perception representation \cite{he2019geonet}, cultural heritage analysis \cite{varriale2022underground}, and medical bone reconstruction\cite{zhang2025pussm}. This will help us make our work more practical.

\section{Conclusion}

In this research, we proposed the ReLPU framework to address the challenge of adapting to the rapid evolution of network structures while ensuring superior performance in point cloud upsampling. ReLPU was designed to be a flexible and adaptable framework that allows the backbone network to be easily replaced with advanced architectures. By enabling the parallel extraction of global and local features, ReLPU significantly enhances the quality and fidelity of upsampled point clouds. 

The ReLPU framework for point cloud upsampling, integrating parallel global and local feature extraction. By employing identical autoencoders for both global and local inputs, our method effectively addresses challenges related to sparsity, noise, and topological inconsistencies. Extensive experiments on the PU1K\cite{qian2021pugcn} and ABC\cite{Koch_2019_CVPR} datasets demonstrated the superiority of ReLPU over state-of-the-art models in terms of geometric fidelity and robustness. Saliency map analysis further validated the importance of combining global and local features for accurate upsampling. The ReLPU framework not only improves the uniformity and detail of point clouds but also enhances its adaptability to real-world applications, making it a versatile solution for future advancements in point cloud processing.


\section{Acknowledge}
This research was supported by the National Natural Science Foundation of China under Grant 61773164.

\bibliographystyle{elsarticle-num} 
\bibliography{reference}


%\begin{thebibliography}{00}


%\bibitem{}

%\end{thebibliography}
\end{document}